\documentclass[sn-mathphys,Numbered]{sn-jnl}

\usepackage{graphicx}%
\usepackage{multirow}%
\usepackage{amsmath,amssymb,mathtools, amsfonts}%
\usepackage{amsthm}%
\usepackage{mathrsfs}%
\usepackage[title]{appendix}%
\usepackage{xcolor}%
\usepackage{textcomp}%
\usepackage{manyfoot}%
\usepackage{booktabs}%
\usepackage{algorithm}%
\usepackage{algorithmicx}%
\usepackage{graphbox} 
\usepackage{algpseudocode}%
\usepackage{graphicx}
\usepackage{listings}%
\usepackage{caption}
\usepackage{subcaption}
\usepackage{enumitem}
\usepackage{orcidlink}

\theoremstyle{thmstyleone}%
\theoremstyle{thmstyletwo}%

\theoremstyle{thmstylethree}%

\begin{document}

\title{Are We Using Autoencoders in a Wrong Way?}


\author[1]{\fnm{Gabriele} \sur{Martino}\orcidlink{0009-0006-3345-1045}}\email{gabriele.martino@isti.cnr.it}

\author[1]{\fnm{Davide} \sur{Moroni}\orcidlink{0000-0002-5175-5126}}\email{davide.moroni@isti.cnr.it}

\author[1]{\fnm{Massimo} \sur{Martinelli}\orcidlink{0000-0001-7419-5099}}\email{massimo.martinelli@isti.cnr.it}

\affil[1]{\orgdiv{Istituto di Scienza e Tecnologie dell’Informazione Alessandro Faedo}, \orgname{CNR}, \city{Pisa}, \postcode{56124}, \state{Italy}}


\abstract{Autoencoders are certainly among the most studied and used Deep Learning models: the idea behind them is to train a model in order to reconstruct the same input data. 
The peculiarity of these models is to compress the information through a bottleneck, creating what is called Latent Space. Autoencoders are generally used for dimensionality reduction,  anomaly detection and feature extraction.
These models have been extensively studied and updated, given their high simplicity and power.
Examples are (i) the Denoising Autoencoder, where the model is trained to reconstruct an image from a noisy one; (ii) Sparse Autoencoder, where the bottleneck is created by a regularization term in the loss function; (iii) Variational Autoencoder, where the latent space is used to generate new consistent data. In this article, we revisited the standard training for the undercomplete Autoencoder modifying the shape of the latent space without using any explicit regularization term in the loss function. We forced the model to reconstruct not the same observation in input, but another one sampled from the same class distribution. We also explored the behaviour of the latent space in the case of reconstruction of a random sample from the whole dataset.}

\keywords{Autoencoder, Neural Network, Deep Learning, Latent Space}

\maketitle

\section{Introduction}

Deep neural networks are at the forefront of Artificial intelligence (AI). They are based on algorithms for learning multiple levels of representation in order to model complex relationships among data. Their design allows them to avoid the over-engineering of the features, leaving the model the responsibility to extract the best information that is useful for the task. The universal approximation theorem \cite{hornik1989multilayer} assesses that shallow neural networks are able to approximate any function, but it is a known fact that deep neural networks (DNNs) perform better than other Machine Learning (ML) methods  \cite{mhaskar2017and}\cite{lee2011unsupervised}\cite{hinton2006reducing} and that this is still a research topic  \cite{sonoda2019transport}. 

Among the most frequently employed and extensively researched DNNs, a prominent category is represented by the AutoEncoders (AEs) family. These models take their success also thanks to their conceptual simplicity.  Their training is accomplished in an unsupervised fashion.
It is only necessary to train the model to create an output identical to the input, after the latter has passed through a bottleneck of the architecture having a smaller dimensionality  \cite{bank2020autoencoders}\cite{zhang2018better}.
The most straightforward architecture of this model family is made by a parameterized function called encoder $ E_\theta(x): \mathbb{R}^n \xrightarrow{} \mathbb{R}^m$, with $ m < n$, $ x \in \mathbb{R}^n$ and parameters $\theta$, and a decoder $ D_\phi(z): \mathbb{R}^m \xrightarrow{} \mathbb{R}^n$,  $ z \in \mathbb{R}^m$ and parameters $\phi$.

This kind of autoencoder is named \textit{undercomplete} and in the case of perfect reconstruction we have $ D_\phi(E_\theta(x)) = x$ \cite{kramer1991nonlinear}.

This model has been widely studied and modified, in particular as regards the mystery around the properties and the power of the Latent Space (LS). The state of the art explores how to make the most of the model also for other tasks. The Denoising Autoencoder \cite{vincent2008extracting} exploits the power of the model showing that it is possible to reconstruct the input data even if it has some additive Gaussian noise. 
Sparse Autoencoders \cite{meng2017research}\cite{makhzani2013k} instead create the bottleneck forcing the activation of a smaller number of nodes in the model or adding a sparsity regularization term for the encoder in the loss function. 
Another interesting model in the family of autoencoders is the Contractive AE (CAE) \cite{rifai2011contractive}. This model also adds a regularization term in the loss for the encoder, but, in this case, trying to encode small variations of the input in small variations in LS. Formally, the regularization term is $ \mathbb{E}[||\nabla_xE_\theta(x))||_{F}^{2}]$ that represents the expectation of Frobenius norms of the variations of the encoded inputs. It is possible to think about the DAE as a CAE, where the small variations in the input are made by Gaussian noise.
Considering the dimensionality reduction task, it has been shown that an autoencoder, with linear or only sigmoid hidden layers, is strongly related to the Principal Component Analysis (PCA) \cite{bourlard1988auto}\cite{chicco2014deep}, and even though the weights of the AE are different, it is possible to recover orthogonal basis using the Singular Value Decomposition (SVD) \cite{plaut2018principal}. Undercomplete AEs are actually more powerful than PCA. They are able to extract the non-linear manifold structure of the input space, whereas the PCA is just a linear projection \cite{hinton2006reducing}.

The LS of these models is still a research topic. Despite the impressive abilities to embed the data manifold into a low-dimensional LS, this representation is usually non-interpretable \cite{leeb2022exploring}. 
However, given the power of the LS, it is conceivable to use the decoder to generate new data from the same encoded distribution. 
Unfortunately, this may lead to poor results, given the irregularity of the space. Regularity is defined as the capability of the space to encode similar data into close space. To solve this issue, Variational Autoencoder (VAE) \cite{kingma2013auto} forces the LS to be encoded in a known distribution, generally Gaussian, allowing a consistent follow-up sampling. This approach received high interest from the research community to better exploit a now regular LS \cite{burgess2018understanding}\cite{higgins2017betavae}\cite{jang2017categorical}\cite{DBLP:conf/nips/SohnLY15}\cite{chen2019isolating}. 

However, all the attempts to explain the LS seem to struggle due to the complexity of the DNNs themself. Moreover, from a manifold learning point of view, also called Non Linear Dimensionality Reduction (NLDR), many of the attempts are made to preserve the topology of the manifold or at least the pairwise distance \cite{lee2007nonlinear}. Known examples are Isomap \cite{tenenbaum2000global} and Locally Linear Embedding (LLE) \cite{roweis2000nonlinear}. In both scenarios, a local neighborhood graph is used to approximate the manifold, which is then used to create a low-dimensional representation that preserves either pairwise geodesic distances (for Isomap) or local linearity of neighborhoods (in LLE). TopoAE \cite{moor2021topological} instead attempts to preserve the topology of the manifold by adding in the loss function a
term that penalizes a topological difference between some information precomputed in the input space and the LS ones. Chart-AEs \cite{schonsheck2020chart} uses multiple overlapping maps for the LS to preserve the geometry of the manifold. Duque et al. \cite{Duque_2020} propose to put as a regularization term in the loss function of an AE the distance with the shape learnt from another manifold learning algorithm that preserves the topology; this allows to exploit the intrinsic invertibility of the decoder. 
It's possible to state that, from this perspective, we almost go in the opposite direction. When we are willing to use NLDR as a feature extraction method for a later classification task, in real cases, the submanifolds of each class can be quite entangled. 
This can cause that learning the topology of the manifold is not enough to get good classification performance\cite{kienitz2022effect}. For this reason, we instead assess that if two class submanifolds information are different, we can ``shrink'' them on themselves in an almost unsupervised fashion to allow a better separability for a later classification, assuming the class information available. More formally, our approach creates a diffeomorphism that helps to extract more separable feature vectors. A similar approach has been used in the Diffeomorphic-AE \cite{bone:hal-01963736} where the deformation of the space is embedded in a regularization term with respect to a specific wanted shape. To allow this behaviour of the latent space we impose the AE to reconstruct in output a random observation sampled from the same distribution of the input observation. This is what we called In-Class distribution Random Sampling Training (ICRST). This idea has some similarities with the Siamese Neural Networks \cite{chicco2021siamese} where two identical networks are trained to have similar embeddings for similar data. However, we focused more on the latent space of the AE, and how this can change. Moreover, in case of the lack of availability of the class information, we can still use our approach in a totally unsupervised fashion. Counterintuitively, we train the AE using a random sample from the dataset to be reconstructed from the input. We called this approach Total Random Sampling Training (TRST). With this later approach, instead, we assume that data could be pushed to rearrange by itself naturally reflecting the similar nature of the observations.

In summary, the main contributions of this article are: 
\begin{enumerate}[label=\roman*]
    \item a novel and simple training framework for Autoencoder models for features extraction;
    \item showing manifold manipulation capabilities of Autoencoders;
    \item showing some information compression results from the natural rearranging of observation in latent space.
\end{enumerate} 
The paper is organized as follows: in Section 2, we report a background on the undercomplete autoencoder; in Section 3, we introduce the In-Class distribution Random Sampling Training (ICRST) method and some implications; in Section 4, we also propose the Total Random Sampling Training (TRST), the extreme case of the ICRST; in Section 5, we show the ICRST from a manifold perspective and some intuitions; Section 6 reports the experimental setup for the ICRST and its results; Section 7 instead reports some results and insights of TRST method; finally Section 8 reports the conclusions and possible future works.
The source code used for the experimentation is available in the following repository: \url{https://github.com/GabMartino/icrst_trst_autoencoder} 

\section{Undercomplete Autoencoder}
In this section, we report the basics formulation of the main Undercomplete Autoencoder-based model.
Let's consider $ X = \{x_1, ..., x_N\} $ where the $ x_i \in \mathbb{R}^n$ are  $N$ observations.

Then we define two general non linear transformations $ z = g(x) $ where $ g(x) : \mathbb{R}^n \xrightarrow{} \mathbb{R}^m$ with $ m < n$ that we'll call encoder, and $ \hat{x} = f(z) $ where $ f(z): \mathbb{R}^m \xrightarrow{} \mathbb{R}^n $ is called decoder. We generally want to minimize the Mean Squared Error (MSE):
\begin{equation}
\begin{split}
    MSE_{AE} &= \frac{1}{2N} \sum_{i}||x_i - f(g(x_i))||^2 \\
            &= \mathbb{E}[(x - f(g(x)))^2]
\end{split}
\end{equation}

Considering that the two functions encoder and decoder are unknown, we can consider this problem as a variational calculus problem, where these two functions depend on some free parameters $ \theta $ and $ \phi$ that need to be optimized. Hence, we could rewrite the problem as follows:
\begin{equation}
    \mathcal{L}(x, \theta, \phi) = \operatorname*{arg\,min}_{\theta,  \phi }\mathbb{E}_x[(x - f(g(x, \theta),\phi))^2]
    \label{equ:classicalAE}
\end{equation}

Now, to find the minimum point of this optimization problem, it would be enough to set the gradient of the two parameters to zero; this is made possible considering that MSE is a convex function. Since the two functions are modelled as neural networks, we can use back-propagation and any Stochastic Gradient Descent method (and its variants) for the optimization.
The minimum point is found when for all the observations $x_i$ in the loss function is zero, that is, the model is able to reconstruct perfectly the input data (it's possible to find a complete computation in Appendix \ref{UndercompleteOptimization}):
\begin{equation}
    f(g(x, \phi),\theta) = x \quad \forall x 
\end{equation}

\section{In-Class Distribution Random Sampling Training}
In this section, we explain our modified training and its implications. \\
Autoencoders are often used as feature extractors for a later classification task. Instead of the total unsupervised training, this information has already been exploited in Charte et al. \cite{charte2021reducing} with three different methods adding regularization terms that include class information in the LS. Similarly, Class-Informed-VAE \cite{nabian2023ci} includes a regularisation term in the Loss function of the vanilla VAE that increases the class distributions' linear separability. Still, another example is Supervised-AE (SAE) \cite{le2018supervised}, which shows higher stability in the training for the classification task when the class label is incorporated in the loss function. Our method includes this information in a simpler way without almost any changes in the loss function, just exploiting an LS deformation that we will later explore.

Let $ X = \{x_1, ..., x_N\} $ where the $x_i \in \mathbb{R}^n$ observation could have been sampled 
from any class $j$ with $j \in {1, ..., M}$ with $M$ number of classes. Let's also define the probability distribution function $p_j(x) $ for each class $j$, we could rewrite the loss function in the Eq. (\ref{equ:classicalAE}) in this way:
\begin{equation}
     \mathcal{L}(x, \theta, \phi) = \operatorname*{arg\,min}_{\theta,  \phi }\mathbb{E}_{x \sim p_j(x)}[(x - f(g(x, \theta),\phi))^2]  \quad \forall j
\end{equation}
Or,
\begin{equation}
     \mathcal{L}(x, \theta, \phi) = \operatorname*{arg\,min}_{\theta,  \phi }\mathbb{E}_{j \in [1,...,M]}[\mathbb{E}_{x \sim p_j(x)}[(x - f(g(x, \theta),\phi))^2]] 
\end{equation}
Where we simply put in evidence the different class distributions. \\

Assuming that all the observations sampled from the same class distribution share similar features, it's possible to force the model to extract only the shared in-distribution features.
So, let's consider $ y \sim p_j(x)$ and $ x \sim p_j(x)$ two independent observations sampled randomly from the same class $j$. We model our loss function accordingly:

\begin{equation}
    \mathcal{L}(x, \theta, \phi) = \operatorname*{arg\,min}_{\theta,  \phi }\mathbb{E}_{y \sim p_j(x), x \sim p_j(x)}[(y - f(g(x, \theta),\phi))^2] \quad \forall j
\end{equation}
Or,
\begin{equation}
    \mathcal{L}(x, \theta, \phi) = \operatorname*{arg\,min}_{\theta,  \phi }\mathbb{E}_{j \in [1,...,M]}[\mathbb{E}_{y \sim p_j(x), x \sim p_j(x)}[(y - f(g(x, \theta),\phi))^2]]
    \label{eq:mainLoss}
\end{equation}

Now it is possible to pose the gradient of loss to zero since it is a convex function to find the minimum, leading to the current results (the complete computation is in Appendix (\ref{AutoOptimiRandom})):
\begin{equation}
\label{eqn:mean_value_convergence}
    \mathbb{E}_{x \sim p_j(x)}[f(g(x,\theta),\phi)] = \boldsymbol{\mu_j} \bigg|_{\mathcal{L}(x, \theta, \phi) = 0} \quad \forall j
\end{equation}
\\
This result is reasonable since we don't make the Autoencoder build the identity function but extract from an observation of a class distribution another observation from the same class, so, in the end, the best guess is the expected value.

From this finding, it's also possible to compute a lower bound for the loss (see Appendix \ref{ReconstErrorRandom}):
\begin{equation}
    \mathcal{L}_j(x, \theta, \phi)\bigg|_{\mathbb{E}[f(g(x,\theta),\phi)] = \boldsymbol{\mu_j}} \geq Var_j(Y) + Var_j( f(g(X, \theta),\phi)) \quad \forall j
\end{equation}
And the Reconstruction Error for the same observation (see Appendix \ref{ReconstErrorRandom}):
\begin{equation}
    \begin{split}
        \mathbb{E}[(f(g(X)) - X)^2] &= \sigma_{f(g(X))}^2 + \sigma_x^2 - 2\mathbb{E}[xf(g(x))] \\
        \quad &= \mathcal{L}_{bound}- 2\mathbb{E}[xf(g(x))] \quad \forall j
    \end{split}
\end{equation}
\\
It's important to notice that, in this case, the reconstruction error is dependent on its own class distribution: the less the variance of that belonging class distribution, the less the reconstruction error.

\section{Total Random Sampling Training}
In this section, we want to show the extreme case of our training method where the sampling is done through the whole dataset, so coming back to totally unsupervised learning.

We remind that our training method is made using the loss function in  Eq. (\ref{eq:mainLoss}), where the sampling happens from the same class distribution. Now, if we consider the whole dataset, we can assume that similar objects should be close in the LS. For this reason, we can relax the conditions of the loss function, letting the model the burden of rearranging the space at best. Hence, let $ x \sim p_x(x), y \sim p_x(x)$ so the two observation are simply randomly sampled from the whole dataset, the loss function becomes:
\begin{equation}
    \mathcal{L}(x, \theta, \phi) = \operatorname*{arg\,min}_{\theta,  \phi }E_{y \sim p(x), x \sim p(x)}[(y - f(g(x, \theta),\phi))^2]
\end{equation}

Note that with this kind of training it is not possible to call this model \textit{Auto}encoder anymore. We'll  explore some insights about this extreme case in Section \ref{TRST_results}.

\section{A manifold learning perspective}
In this section, we report some important notions on Manifold learning related to Autoencoders and some insights on similarities with the Denoising autoencoder.\\
Many machine learning algorithms exploit the idea that data concentrates around a lower-dimensional manifold or a small set of such manifolds. Autoencoders take this idea further and aim to learn the structure of the manifold \cite{Goodfellow-et-al-2016} \cite{shi2021discussion}.
 A characterization of a manifold is the set of its tangent planes. At a point \textbf{x} on a \textit{d}-dimensional manifold, the tangent plane is given by $d$ basis vectors that span the local directions of variation allowed on the manifold.

The important principle is that the autoencoder can afford to represent only the variations that are needed to reconstruct the training examples \cite{Goodfellow-et-al-2016}. This well explains the irregularity of the LS and why it is not easy to use the whole lower-dimensional space to generate new coherent data as a VAE does (at least around the fixed prior distribution) \cite{bengio2013representation}.

Following this line, a Denoising autoencoder learns to reconstruct a data point from its perturbation from the manifold. We remind that a DAE minimize the vector $ (f(g(\Tilde{\boldsymbol{x}})) -\boldsymbol{x})$ where $ \Tilde{\boldsymbol{x}} =\boldsymbol{x} + \boldsymbol{\epsilon} $ where $ \boldsymbol{\epsilon} \sim N(0, \sigma^2 I)$, multivariate standard distribution with zero mean and $\sigma^2 I$ as variance.

\begin{figure}[ht]
\centering
\begin{subfigure}{.5\textwidth}
  \centering
  
\includegraphics[width=7cm]{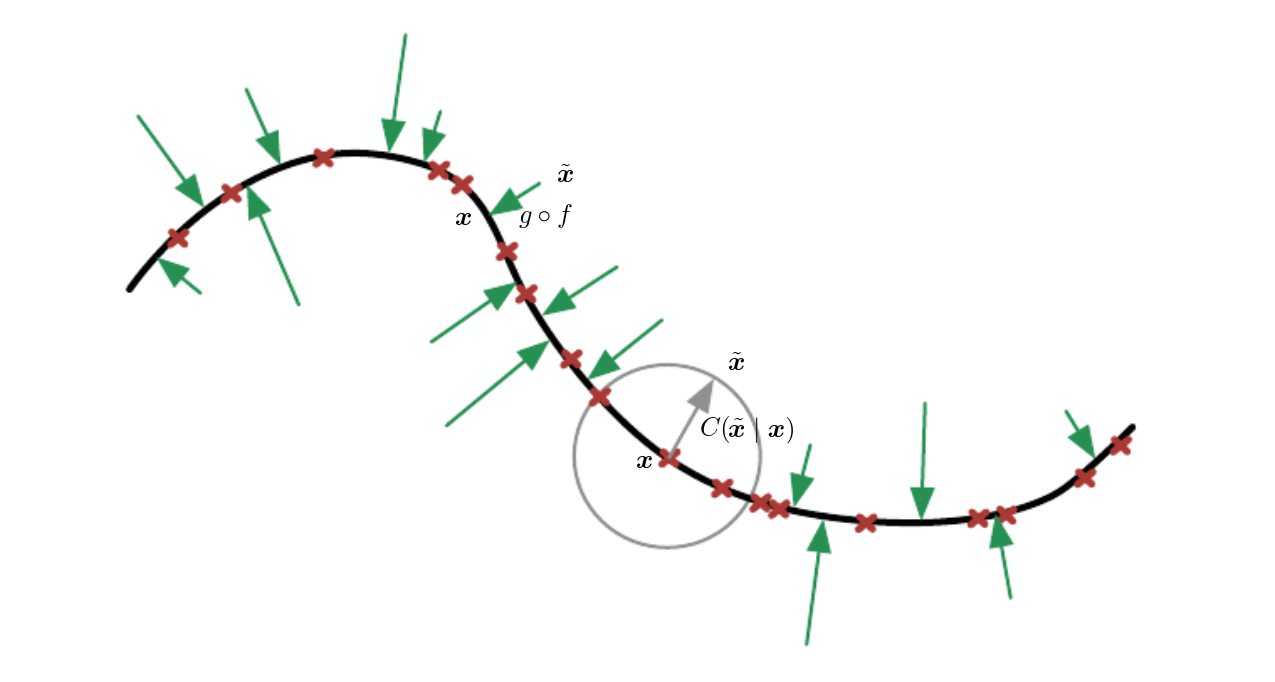}
\caption{}
  \label{DAE_manifold}
\end{subfigure}%
\begin{subfigure}{.5\textwidth}
  \centering
  \includegraphics[width=7cm, trim={0 0 0 3cm}, clip]{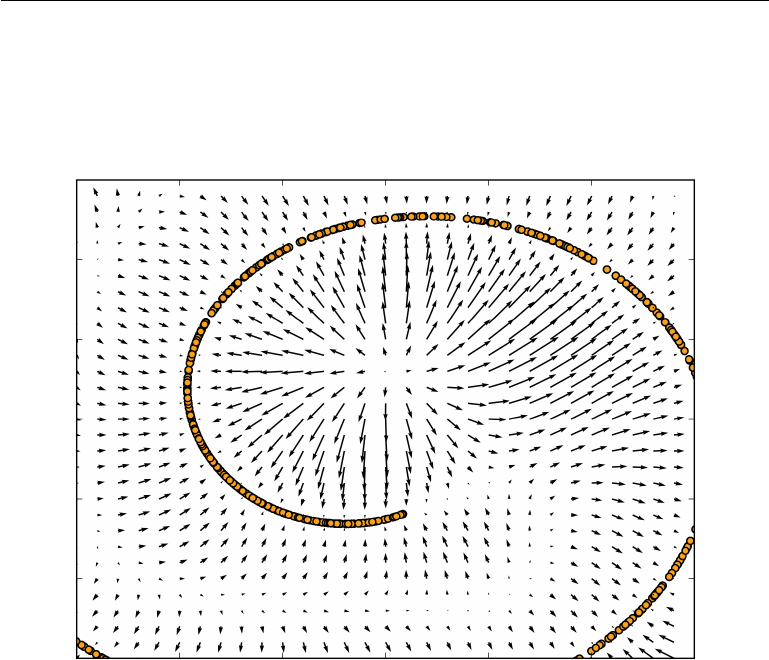}
\caption{ }
  \label{vectorFIeld}
\end{subfigure}\\
\caption{(a) DAE manifold learning representation. (b) DAE manifold learning vector field.}
\end{figure}
\noindent
Fig. \ref{DAE_manifold} shows a visual representation of the DAE training on the manifold illustrated with a bold black line. The grey arrow and circle represent the corruption process $ C(\Tilde{\boldsymbol{x}} | \boldsymbol{x}) = N(\mu = \boldsymbol{x}, \Sigma = \sigma^2 I)$. The green arrows represent the vector field $f(g(\boldsymbol{x})) - \boldsymbol{x}$. The vector $f(g(\Tilde{\boldsymbol{x}})) - \Tilde{\boldsymbol{x}}$ points approximately toward the nearest point on the manifold.
If we would draw the vector field created from this learning method, it'd seem something like what is shown in Fig. \ref{vectorFIeld}, where any new sample introduced to the model falls into the closest point on the manifold.

Following the same conceptual manifold representation for DAE, we could think of our approach as an extreme case, where the "corruption" process is not due to a Gaussian noise but is represented by another sample from the same manifold. 

\begin{figure}[H]
\captionsetup{justification=centering}
\includegraphics[width=10cm]{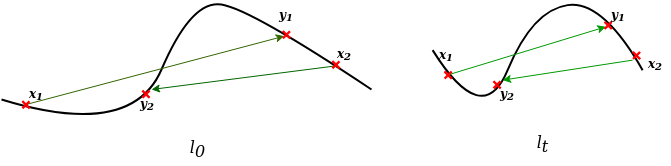}
\centering
\caption{Manifold learning compression process after $t$ steps of training.}
    \label{DAE_manifold_1}
\end{figure}

This brings the manifold to be always more shrunk along the layers of the deep neural network as shown in Fig. \ref{DAE_manifold_1}. This visualization also well explains the findings at the equation (\ref{eqn:mean_value_convergence}) where the observations collapse around the mean value of the manifold.

\noindent

It's possible to rearrange the Eq. \ref{eq:mainLoss} of our training method as a regularized form of the classical MSE for the undercomplete autoencoder. We can first sum and subtract $ 2\mathbb{E}[xf(g(x))]$ and then we can also consider that the second order momentum of $X$ and $Y$ is the same since they are the same random variable, resulting in:


\begin{multline}
     \mathcal{L}(x, \theta, \phi) = \operatorname*{arg\,min}_{\theta,  \phi } \biggl[ \mathbb{E}_{j \in [1,...,M]}\bigl[\mathbb{E}_{p_j(x)}[(x - f(g(x, \theta),\phi))^2] +  \\  + 2\mathbb{E}_{p_j(x)}[f(g(x, \theta),\phi))( x - \mathbb{E}_{p_j(x)}[X])]\bigr]     \biggr]
\end{multline}

Now it is possible to visualize a penalization term for the distance between the observation and its respective distribution expected value.
This behaviour brings the model not only to learn the manifold shape, so a lower-dimensional representation, but also to compress it in the latent space.
\begin{figure}[H]
\centering
\begin{subfigure}{.5\textwidth}
  \centering
\includegraphics[width=6cm]{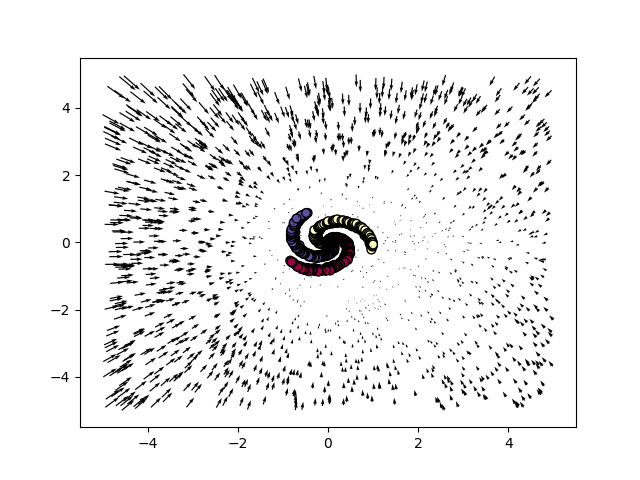}
\caption{}
  \label{DAE_sink_far}
\end{subfigure}%
\begin{subfigure}{.5\textwidth}
  \centering
  \includegraphics[width=6cm, trim={0 0 0 0}, clip]{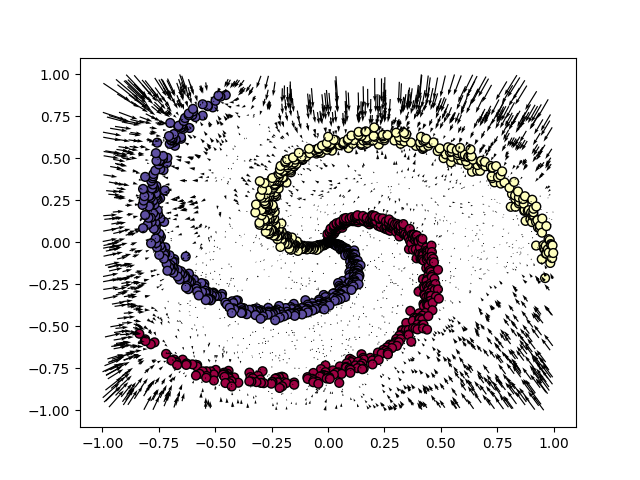}
\caption{ }
  \label{DAE_attractor}
\end{subfigure}\\
\caption{ DAE's vector field (a) from far acts as sink (b) from close}
\label{DAE}
\end{figure}
\begin{figure}[H]
\centering
\begin{subfigure}{.5\textwidth}
  \centering
  
\includegraphics[width=7cm]{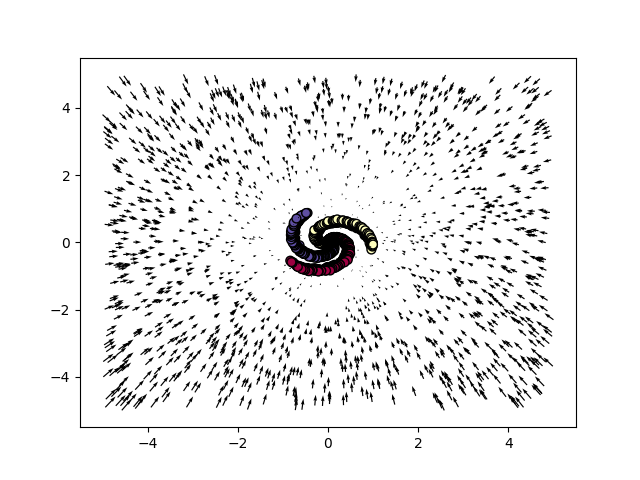}
\caption{}
  \label{In_class_sink}
\end{subfigure}%
\begin{subfigure}{.5\textwidth}
  \centering
  \includegraphics[width=7cm, trim={0 0 0 0}, clip]{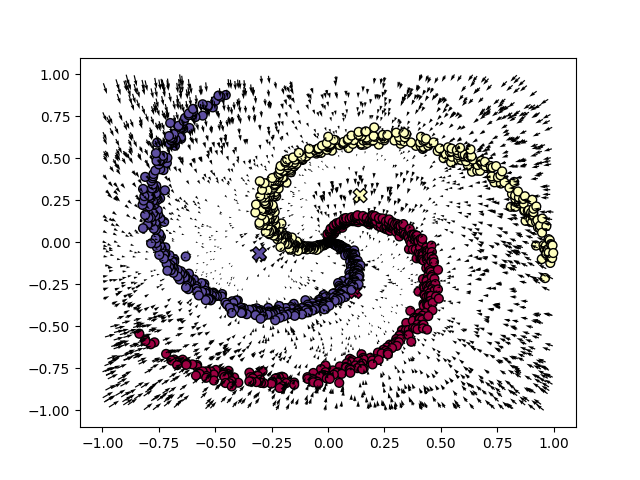}
\caption{ }
  \label{in_class_sink_close}
\end{subfigure}\\
\caption{In-class distribution trained AE (a) from far (b) from close}
\label{inclassAE}
\end{figure}

In the Fig. \ref{DAE} and Fig. \ref{inclassAE}, we repropose the same visualization in Fig. \ref{vectorFIeld} with a toy dataset and a toy model to show the differences in the distortion of the space. As it is possible to see, from afar the two models act in almost equal ways. Closer to the data instead, the in-class distribution trained model acts as we found where the manifold is dragged towards the mean value.
In support of our method, Vincent et al. \cite{vincent2008extracting} proposed to use stacked DAE as initialization for any Deep Neural Network to create more robust features.

\section{ICRST Experimental Setup and Classification Performance}\label{ICRST_results}
In this section, we report the experimental pipeline of the In-Class distribution Random Sampling training method (ICRST) and the results.
We used several models and datasets shown in tab. \ref{tab:Tab1}). For the CNNs models, we extracted the features from the last features map of the encoder and computed the Average Global Pooling, to handle fewer dimensions. All the image datasets are first normalized in the range $[0,\, 1]$ and then standardized per channel. The non-image dataset is just normalized.

Now, to test how our training method affects the shape of the LS, we gradually `inject' it into the standard method. To accomplish this, we defined a Bernullian random variable $ \boldsymbol{q} \sim \mathbb{B}(\boldsymbol{p})$ with $ 0 \leq \boldsymbol{p} \leq 1$ as hyperparameter. This means that $ \boldsymbol{q} = 1$ with probability $\boldsymbol{p}$ and $ \boldsymbol{q} = 0$ with probability $  \boldsymbol{1 - p}$. We set the experiments in such a way that if the random variable $ \boldsymbol{q} = 0 $, the training will be set in `standard mode' (the model should match the same data in input), instead if $ \boldsymbol{q} = 1$ the training will shift to ICRS. This sampling is made at every step of the training. This means that if $ p = 0.2$, the $20\%$ of the times the image for the reconstruction error computation is sampled randomly from the same class distribution. This approach allows us to understand how much our method affects the LS topology and, also, if training could benefit just partially from our training as a `fine tuning' method.
The models are developed using Pytorch Lightning \cite{NEURIPS2019_9015} \cite{falcon2019pytorch}, batch size of 512 (64 for the CNN\textsubscript{2}), learning rate of $10*10^{-4}$, 50 epochs.
After the training phase, we used the features extracted from the LS for a classification task using several classifiers: SVM (with RBF), Random Forest, MLP, Gaussian Naive Bayes with SKLearn library \cite{scikit-learn}. All the classifiers are used with the default hyperparameters. The datasets used are: MNIST \cite{deng2012mnist}, Fashion-MNIST \cite{xiao2017fashionmnist}, CIFAR-10 \cite{cifar_10}, Caltech101 \cite{li_andreeto_ranzato_perona_2022},  BreastCancer \cite{misc_breast_cancer_wisconsin}.
\begin{figure}[H]
\centering
\begin{subfigure}{.5\textwidth}
  \centering
  \includegraphics[width=1.1\linewidth, trim={0 0 0 1.45cm},clip]{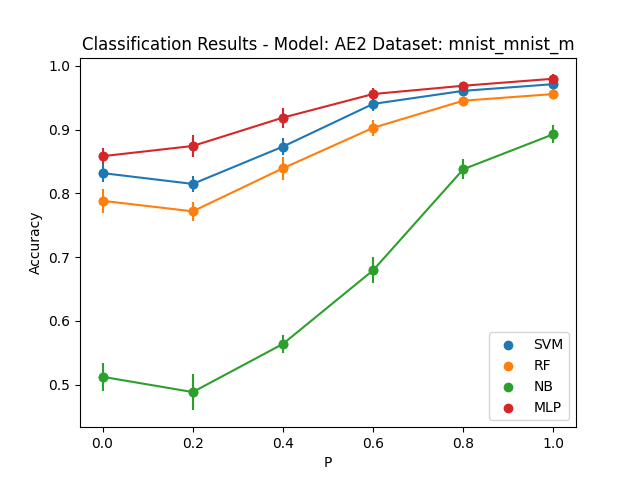}
  \caption{MNIST}
\end{subfigure}%
\begin{subfigure}{.5\textwidth}
  \centering
  \includegraphics[width=1.1\linewidth, trim={0 0 0 1.45cm},clip]{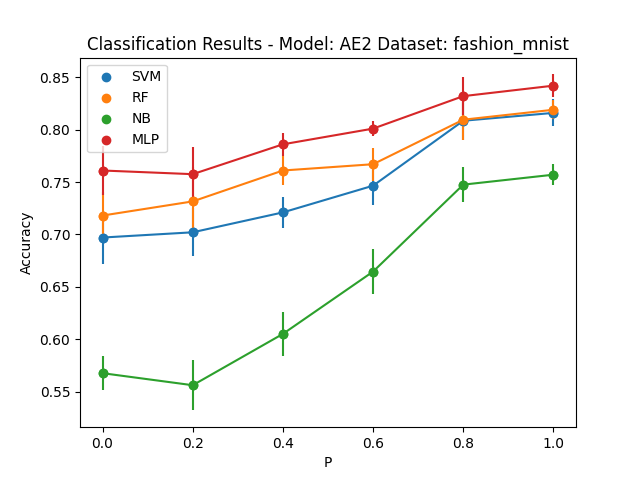}
  \caption{Fashion-MNIST}
\end{subfigure}\\
\begin{subfigure}{.5\textwidth}
  \centering
  \includegraphics[width=1.1\linewidth, trim={0 0 0 2cm},clip]{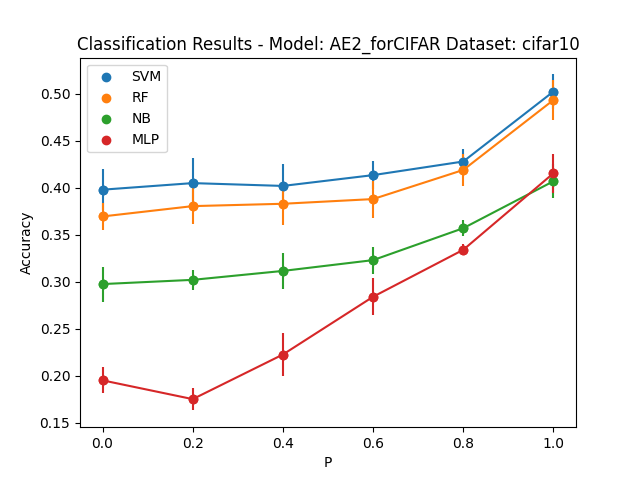}
  \caption{CIFAR10}
\end{subfigure}%
\begin{subfigure}{.5\textwidth}
  \centering
  \includegraphics[width=1.1\linewidth, trim={0 0 0 1.45cm},clip]{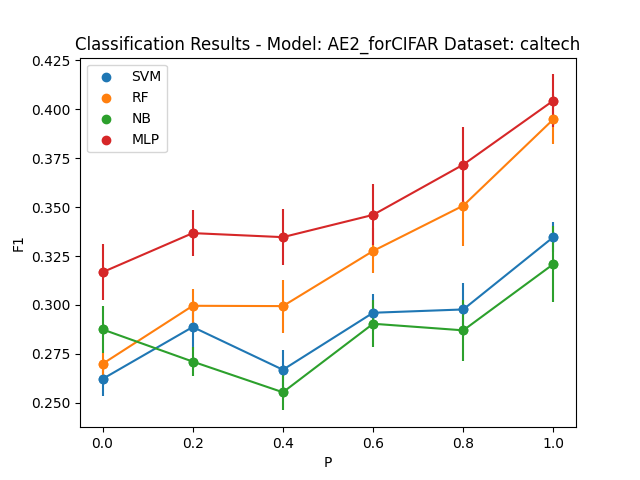}
  \caption{Caltech101}
\end{subfigure}\\
\begin{subfigure}{.5\textwidth}
  \centering
  \includegraphics[width=1.1\linewidth, trim={0 0 0 1.45cm},clip]{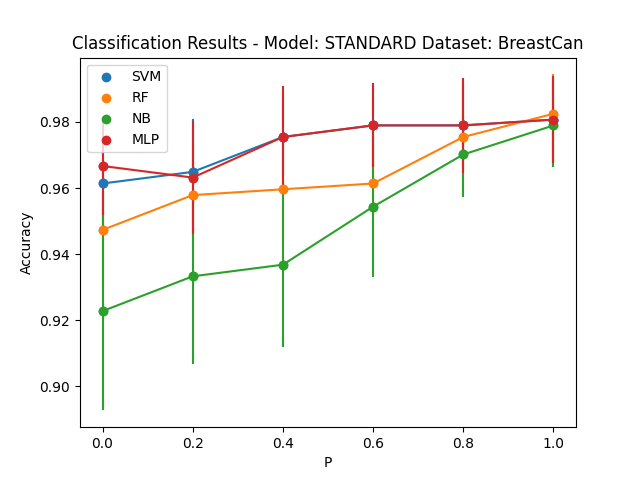}
  \caption{Breast Cancer}
\end{subfigure}

\caption{Classification Accuracy/F1 Score results for different classifiers (SVM, RF, NB, MLP) for several values of p}
\label{fig:ClassResults}
\end{figure}

\begin{table*}
    \centering
    \renewcommand{\arraystretch}{1.1}
    \begin{tabular}{l ccccc}
        \toprule
        Method  & MNIST & F-MNIST  & CIFAR10  & Caltech101 & BreastCancer \\
        \midrule
        Standard & $0.85 \pm 0.02$& $0.76 \pm 0.02$ & $0.19 \pm 0.01$ & $0.32 \pm 0.01$ & $0.96 \pm 0.02$\\
        ICRST & $\mathbf{0.97 \pm 0.01}$ & $ \mathbf{0.84 \pm 0.01}$ & $\mathbf{0.41 \pm 0.02}$& $\mathbf{0.40 \pm 0.01}$ & $\mathbf{0.97 \pm 0.02}$\\
        \bottomrule
    \end{tabular}
    \caption{Classification performance for MLP classifier}
    \label{tab:performance_table_1}
\end{table*}
Fig. \ref{fig:ClassResults} reports the Accuracy or F1 score results for all the models, for all the datasets and several values of the hyperparameter $p$ of the Bernullian random variable. The 95\% percentile confidence intervals are computed after 10-cross-fold-validation.
We remind that when $p = 1.0$, the training is purely performed in ICRS mode, while conversely, when $p = 0.0 $, the autoencoder is trained in the standard way.  

As can be seen, the accuracy significantly improves in all the cases. A first intuition of this behaviour can be that the subspaces of each class distribution are reorganized, so we have a better disentanglement between them. This, of course, increases the quality of the features extracted. These results are summarized in Tab. \ref{tab:performance_table_1}. To be noticed that for Caltech101 Dataset the F1 score is reported instead of the Accuracy given the unbalanced nature of the data. 

\begin{figure}[H]
\centering
\begin{subfigure}{0.32\textwidth}
  \includegraphics[width=\linewidth, trim={0 0 0 2cm},clip]{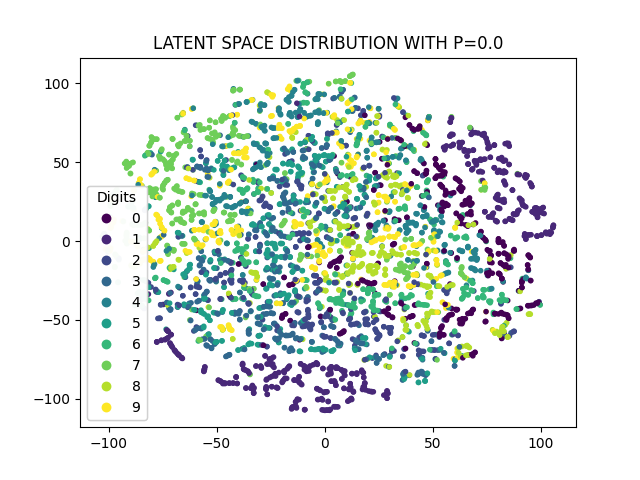}
  \caption{P = 0.0}
  \label{fig:sub1}
\end{subfigure}
\hfill
\begin{subfigure}{0.32\textwidth}
  \includegraphics[width=\linewidth, trim={0 0 0 2cm},clip]{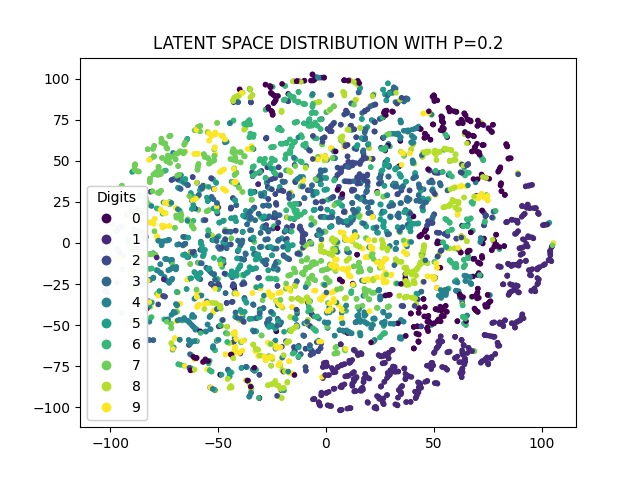}
  \caption{P = 0.2}
\end{subfigure}
\begin{subfigure}{0.32\textwidth}
  \includegraphics[width=\linewidth, trim={0 0 0 2cm},clip]{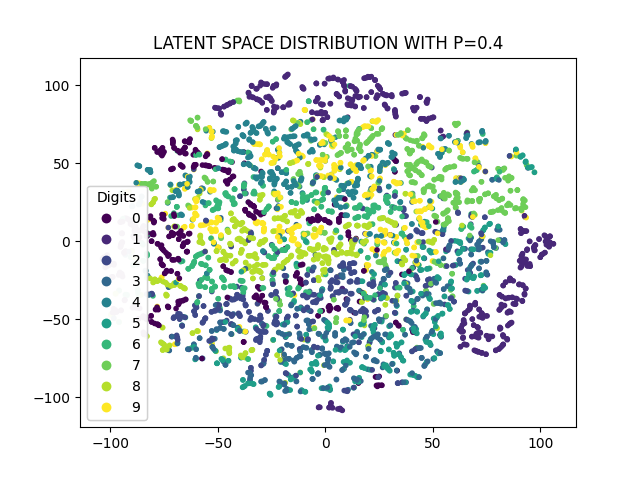}
  \caption{P = 0.4}
\end{subfigure}
\begin{subfigure}{0.32\textwidth}
  \includegraphics[width=\linewidth, trim={0 0 0 2cm},clip]{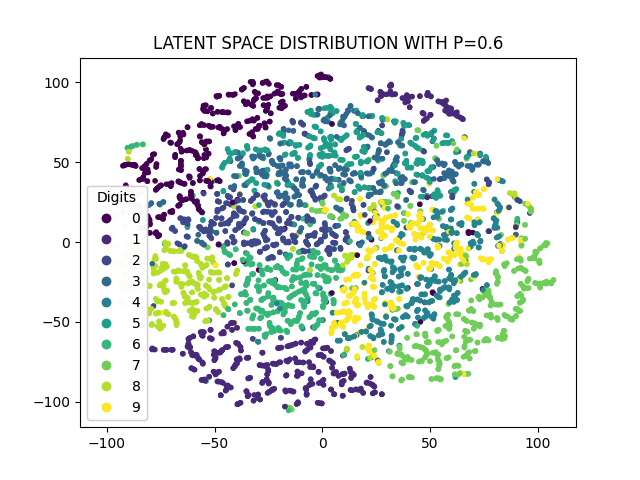}
  \caption{P = 0.6}
\end{subfigure}
\begin{subfigure}{0.32\textwidth}
  \includegraphics[width=\linewidth, trim={0 0 0 2cm},clip]{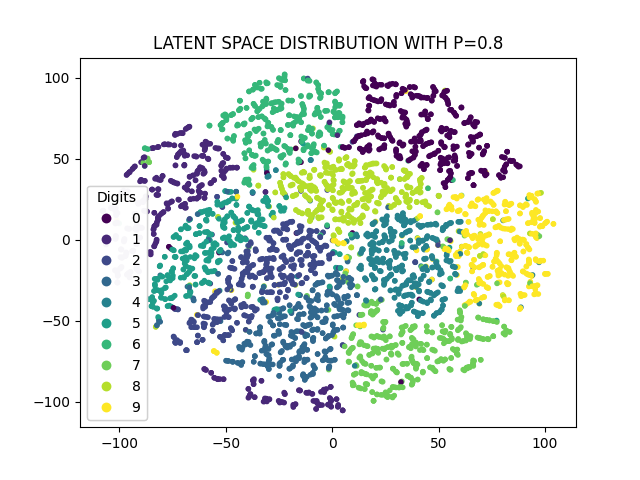}
  \caption{P = 0.8}
\end{subfigure}
\begin{subfigure}{0.32\textwidth}
  \includegraphics[width=\linewidth, trim={0 0 0 2cm},clip]{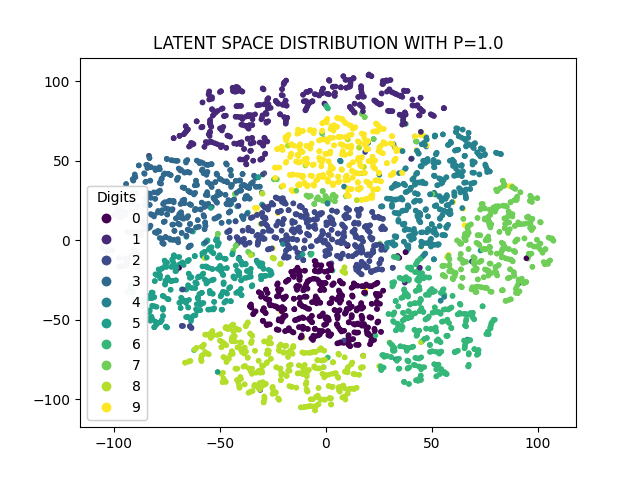}
  \caption{P = 1.0}
\end{subfigure}
\caption{t-SNE projection of the LS for MNIST dataset with in-class distribution training.}
\label{fig:tsne_mnist}
\end{figure}

To better visualize the shape of LS, we used a 2-dimensional t-SNE projection, shown in Fig. \ref{fig:tsne_mnist}, where we plotted the LS of the MNIST dataset. The LS regularize by itself without any additional regularity term in the loss function (like VAE-like models do); this could well explain the higher separability of the class distribution and, thus, the improvements in the classification accuracy.

\section{On Encoding of Information via Natural Rearranging with TRST}\label{TRST_results}
In this section, we report some insights about the Total Random Sampling Training (TRST) and how this will affect the manifold shape and its information encoding capabilities, plus some results on classification performance.

Learning disentangled representation in data can be useful for a large variety of tasks and domains. It is possible to define a disentangled representation as one where single latent units are sensitive to changes in single generative factors while being relatively invariant to changes in other factors \cite{higgins2017beta}\cite{bengio2013representation}\cite{ridgeway2016survey}. Examples of factors are: position, scale, lightning or colour, rotation etc. This could allow a higher explainability of deep learning reasoning. However, the assumption that the data should rearrange themselves in an unsupervised way following simple generative factors is actually quite strong and fundamentally impossible without inductive bias on both the model and the data \cite{locatello2019challenging}.
InfoGAN \cite{chen2016infogan} tries to reach an interpretable representation by adding a regularization term that maximizes the Mutual Information (MI) between the latent codes and the represented data. $\beta$-VAE \cite{burgess2018understanding} achieves the disentanglement by adding the hyperparameter $\beta$ to the Kullback-Leibler (KL) divergence term of the loss. MI (and its relation with the KL divergence) is actually largely used to compute non linear relationships between two random variables $\boldsymbol{x}$ and $\boldsymbol{y}$ \cite{qian2019enhancing}\cite{phuong2018mutual}\cite{ye2021learning}\cite{rodriguez2021disentanglement}. Given the probabilistic nature of these metrics, generally all the generative models develop their loss function starting from log-likelihood respect to the dataset: $\log p(x)$. 
However, our training methods remain on the ``frequentist approach'', showing similar results of ``probabilistic based models'' with the advantage of high ease in feature handling. We remember that the majority of VAE-like models use the encoder to predict the mean and variance of the Gaussian distribution to allow the training by the use of the reparametrization trick; then we sample from that distribution for the reconstruction, that is on due of the decoder. This brings the VAE models to be not particularly suitable for feature extraction. In fact, it is possible to consider the TRST as a VAE where the sampling is made at the output and not in the latent space. Moreover, it is also possible to find some similarities between the CI-VAE \cite{nabian2022ci}, where the class information is embedded in a VAE via a small NN in parallel at the decoder and our ICRST.

\begin{figure}[ht]
\centering
\begin{subfigure}{.5\textwidth}
  \centering
  \includegraphics[width=1.1\linewidth, trim={0 0 0 2cm },clip]{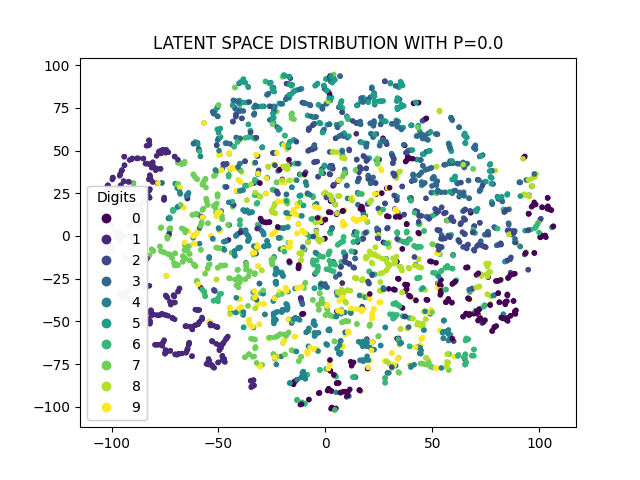}
  \caption{Standard}
\end{subfigure}%
\begin{subfigure}{.5\textwidth}
  \centering
  \includegraphics[width=1.1\linewidth, trim={0 0 0 2cm },clip]{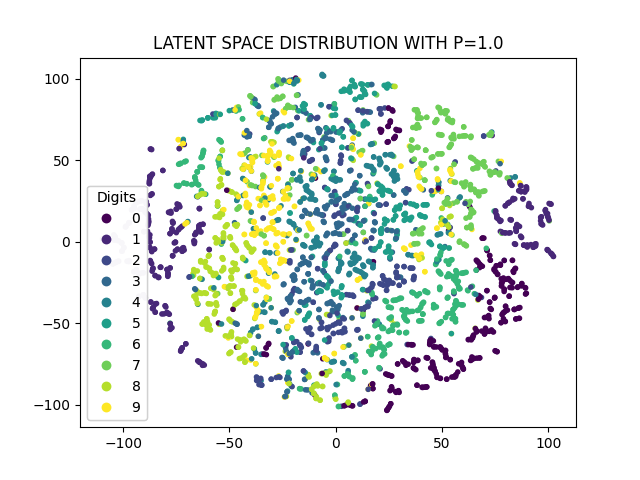}
  \caption{TRST}
\end{subfigure}
\caption{LS Distribution in t-SNE for Total Random Sampling training for MNIST.}

\label{total_random_distribution}
\end{figure}

We believe that TRST leads to a natural rearranging of the latent codes, compressing the information where similar data are closer to each other. Intuitively, when in the loss function (e.g. MSE), we sample two observations of the dataset that are similar (but not the same), the model tends to encode them closely. Instead, when the two observations are very different, the model struggles to reconstruct one observation from the other. The difference between ICRST and TRST with a classification dataset is that ICRST, in some way, forces the model to shrink same-class observation embeddings regardless of the similarity with other class distributions. Indeed, Fig.(\ref{total_random_distribution}) shows a comparison of the topology of the latent space built from a classical AE training and a TRST of the MNIST dataset; an overlapping of the distributions here is totally plausible. As an example, considering a distorted observation of the digit ``7'' could be closer to the distribution of the digit "1" with respect to the mean value of the distribution of ``7''s itself.
For this reason, we can suppose that TRST increases the MI in the neighborhood, possibly later used for learning disentangled representations.

To empirically prove this, Tab. \ref{tab:Mi_table} shows a significant increase in the MI of the latent space for MNIST and Fashion-MNIST datasets, but not for CIFAR10 and Caltech101. To compute the MI, we proceeded as follows: at first, we sampled $N=150$ observations from the latent space and extracted the KNN embeddings for each sample, using Euclidean Distance, with $K=20$; then we computed the according MI between the samples and their K-NN data observations from the dataset. The MI is computed as the average of the MI for each channel of the images.
It's interesting to note that, for the CIFAR10 and Caltech101 dataset, this increase in the MI is not shown. This is likely due to the nature of the data itself. Same class observations could be quite different, with similarities not ``detected'' from the MI computation. 

\begin{table*}[ht]
    \centering
    \renewcommand{\arraystretch}{1.1}
    \begin{tabular}{l cccc}
        \toprule
        Method & MNIST & Fashion-MNIST  & CIFAR10 & Caltech101\\
        \midrule
        MI\textsubscript{Standard} & $0.40 \pm 0.01$& $0.40 \pm 0.01$ & $0.25 \pm 0.00$ & $0.16 \pm 0.00$ \\
        MI\textsubscript{TRST} &  $\mathbf{0.45 \pm 0.01}$ & $ \mathbf{0.44 \pm 0.01}$ & $ 0.25 \pm 0.00$ & $0.16 \pm 0.00$\\
        \bottomrule
    \end{tabular}
    \caption{Mutual information}
    \label{tab:Mi_table}
\end{table*}

In Tab. \ref{tab:performance_table_TRST}, we show some results that follow what we found in Tab. \ref{tab:Mi_table}. The table reports the MLP classification performance using the features vectors extracted from the AEs latent space. We compared the performance between Standard training for AE and the TRST, following the training process and hyperparameters described in Section 6. It's possible to note that these results reflect in some way the ones in Tab. \ref{tab:Mi_table}.
Intrinsically similar class instances (like in MNIST and Fashion-MNIST datasets) take advantage of TRST, given that it can naturally rearrange their embeddings in latent space accordingly. For more complex datasets instead, where we expect a more sparse representation manifold, the model struggles to place near observations that should represent the same class, given that belonging to that class could depend on some hidden and hard-to-find latent features. However, we believe that, with a more ``aggressive'' bottleneck, these results will be visible in complex data as well, but we will leave these experiments for future works.

\begin{table*}[ht]
    \centering
    \renewcommand{\arraystretch}{1.1}
    \begin{tabular}{l ccccc}
        \toprule
        Method  & MNIST & F-MNIST  & CIFAR10  & Caltech101 & BreastCancer \\
        \midrule
        Standard & $0.86 \pm 0.02$& $0.75 \pm 0.02$ & $\mathbf{0.17 \pm 0.01}$ & $\mathbf{0.33 \pm 0.02}$ & $0.96 \pm 0.03$\\
        TRST &  $\mathbf{0.90 \pm 0.02}$ & $ \mathbf{0.89 \pm 0.00}$ & $ 0.09 \pm 0.01$& $0.17 \pm 0.01$& $\mathbf{0.98 \pm 0.02}$\\
        \bottomrule
    \end{tabular}
    \caption{Classification performance for MLP classifier}
    \label{tab:performance_table_TRST}
\end{table*}

\section{Conclusions and Future Works}
Non-linear dimensionality reduction methods, also known as manifold learning, are important techniques used to extract the most essential and minimal information from data. These methods reveal that the topology of the manifold on which the data lie on is extremely important, because manifolds encode information about the transformation between one point representation in the space to another. We have shown that it is possible to exploit the versatility of the AEs to deform these manifolds for our purpose (like in the ICRST) or to reveal important data information (like in the TRST). 
In ICRST we have shown that, with almost no changes in the training method, it is possible to achieve better results in the case we aim to use the AE's encoder as a features extractor. Moreover,  we also achieve better classification results when we have high mutual information shared in the class distribution in TRST, where the training is made completely random, going in totally opposite direction with respect to the main reasoning of AEs. Given the higher complexity of the loss function in these cases, a smaller batch size for the training is suggested.

We believe that ICRST could have important results in Unsupervised Domain Adaptation as well. Considering that we use an AE model to reconstruct a different observation but from the same class distribution, this brings the model to create a more robust feature that abstracts better the class information, avoiding encoding relevant information like the domain ones. We already observed these results in preliminary experiments, at least in datasets where the in-class observations share some information (have high mutual information). However, we are going to leave this discussion for future work.

The similarities between a VAE and the TRST are direct, and this brings to a possible common mathematical description that also involves mutual information, as found in our results. This paper wants to report some results that could lead to a better comprehension of these models of how the data information is stored and encoded, giving a common vision of all the modern VAE-like models.

\begin{appendices}

\section{Undercomplete Autoencoder Optimization}\label{UndercompleteOptimization}

In this section we report the computations for the Undercomplete Autoencoder optimization that will be necessary to understand the differences with our training method.
Let $ X = \{x_1, ..., x_N\} $ where the $ x_i \in \mathbb{R}^n$ are  $N$ observations. Let $ z = g(x) $ where $ g(x) : \mathbb{R}^n \xrightarrow{} \mathbb{R}^m$ with $ m < n$ and $ \hat{x} = f(z) $ where $ f(z): \mathbb{R}^m \xrightarrow{} \mathbb{R}^n $. 
Then we model the Loss function with the Mean Squared Error (MSE):
\begin{equation}
    \mathcal{L}(x, \theta, \phi) = \operatorname*{arg\,min}_{\theta,  \phi }\mathbb{E}[(x - f(g(x, \phi), \theta))^2]
\end{equation}

To optimize this function is enough to set the gradient to zero in view of the convexity. Note that, of course, it is not necessary to compute the gradient to visualize that: 

\begin{equation}
\begin{split}
    \nabla_{\theta, \phi} \mathcal{L}(x, \theta, \phi) &= 0 \\
    \nabla_{\theta, \phi}  \mathbb{E}[(x - f(g(x, \phi), \theta))^2] &= 0  \\
      \mathbb{E}[\nabla_{\theta, \phi}(x - f(g(x, \phi), \theta))^2] &= 0 \\
      \mathbb{E}[\nabla_{\theta, \phi}(x^2 - 2xf(g(x, \phi), \theta) + f(g(x, \phi), \theta)^2)] &= 0 \\
      \mathbb{E}[-2x\nabla_{\theta, \phi}f(g(x, \phi), \theta) + \nabla_{\theta, \phi} f(g(x, \phi), \theta)^2] &= 0 \\
      \mathbb{E}[ \nabla_{\theta, \phi} f(g(x, \phi), \theta)^2] &= \mathbb{E}[2x\nabla_{\theta, \phi}f(g(x, \phi), \theta)]
      \label{eq:underOptim}
\end{split}
\end{equation}

To simplify the notation, let's pose $z = g(x, \phi)$ for the moment.

So,
\begin{equation}
      \nabla_{\theta}f(z, \theta) =  \frac{ \partial f(z, \theta)}{\partial \theta}, \nabla_{\phi}f(z, \theta) =  \frac{ \partial f(z, \theta)}{\partial \phi} = \frac{ \partial f(z, \theta)}{\partial z}\frac{ \partial z}{\partial \phi}
\end{equation}

Then,
\begin{equation}
\label{eq:a4}
    \nabla_{\theta, \phi} f(g(x, \phi), \theta) = \frac{ \partial f(g(x, \phi), \theta)}{\partial \theta} +  \frac{ \partial f(g(x, \phi), \theta)}{\partial g(x, \phi)}\frac{ \partial g(x, \phi)}{\partial \phi}
\end{equation}
and,
\begin{align}
\label{eq:a5}
    \nabla_{\theta, \phi} f(g(x, \phi), \theta)^2 &= 2f(g(x, \phi), \theta)\left(\frac{ \partial f(g(x, \phi), \theta)}{\partial \theta} + \frac{ \partial f(g(x, \phi), \theta)}{\partial g(x, \phi)}\frac{ \partial g(x, \phi)}{\partial \phi}\right)
\end{align}

Finally, inserting eq. \ref{eq:a4} and eq. \ref{eq:a5} into  eq. \ref{eq:underOptim}, we get:
\begin{equation}
\begin{split}
    \mathbb{E}\biggl[f(g(x, \phi), \theta)\left(\frac{ \partial f(g(x, \phi), \theta)}{\partial \theta} + \frac{ \partial f(g(x, \phi), \theta)}{\partial g(x, \phi)}\frac{ \partial g(x, \phi)}{\partial \phi}\right)\biggr] &= \mathbb{E}\biggl[x\biggl( \frac{ \partial f(g(x, \phi), \theta)}{\partial \theta} + \\
    \quad &\frac{ \partial f(g(x, \phi), \theta)}{\partial g(x, \phi)}\frac{ \partial g(x, \phi)}{\partial \phi}\biggr)\biggr]
\end{split}
\end{equation}
\noindent
If we call $\frac{ \partial f(g(x, \phi), \theta)}{\partial \theta} +  \frac{ \partial f(g(x, \phi), \theta)}{\partial g(x, \phi)}\frac{ \partial g(x, \phi)}{\partial \phi} = \boldsymbol{h}$, then:

\begin{equation}
     \mathbb{E}[f(g(x, \phi), \theta)\boldsymbol{h}] =  \mathbb{E}[x\boldsymbol{h}]
    \label{undetachable}
\end{equation}
And as expected, the equality holds when:
\begin{equation}
    f(g(x, \phi),\theta) = x
\end{equation}
\noindent
It's important to note for our training method that ,in the equation \ref{undetachable}, $x $ and $ \boldsymbol{h}$ are dependent, so it is not possible to split the expected value.

\section{Undercomplete Autoencoder Optimization with Random Sampling}\label{AutoOptimiRandom}
Here were will follow the same approach of the vanilla undercomplete autoencoder training, focusing only on the differences. At first, we will report the relaxed Total Random Sampling Training (TRST) method; then we'll see the In-Class Distribution Random Sampling (ICRST).
\begin{equation}
    \mathcal{L}(x, \theta, \phi) = \operatorname*{arg\,min}_{\theta,  \phi } \mathbb{E}\left[(y - f(g(x, \phi), \theta))^2\right]
\end{equation}
Where $ y \sim p(x), x \sim p(x)$, x and y are extracted from the same distribution but are independent. Trying to solve this objective function we have:

\begin{equation}
       \mathbb{E}[ \nabla_{\theta, \phi} f(g(x, \phi), \theta)^2] =  \mathbb{E}[2y\nabla_{\theta, \phi}f(g(x, \phi), \theta)]
\end{equation}

Given the independence of the two random variables.
\begin{equation}
\begin{split}
       \mathbb{E}[ \nabla_{\theta, \phi} f(g(x, \phi), \theta)^2] &=  \mathbb{E}[2y] \mathbb{E}[\nabla_{\theta, \phi}f(g(x, \phi), \theta)] \\
       \mathbb{E}[ \nabla_{\theta, \phi} f(g(x, \phi), \theta)^2] &= 2\boldsymbol{\mu} \mathbb{E}[\nabla_{\theta, \phi}f(g(x, \phi), \theta)] \\
       \mathbb{E}[f(g(x, \phi), \theta)\boldsymbol{h}] &= \boldsymbol{\mu}\mathbb{E}[\boldsymbol{h}] \\
       \frac{\mathbb{E}[f(g(x, \phi), \theta)\boldsymbol{h}]}{\mathbb{E}[\boldsymbol{h}] } &= \boldsymbol{\mu} \\
        \mathbb{E}[ \frac{f(g(x, \phi), \theta)\boldsymbol{h}}{\boldsymbol{h}}] &= \boldsymbol{\mu} \\
        \mathbb{E}[f(g(x, \phi), \theta) ] &= \boldsymbol{\mu}
\end{split}
\end{equation}

Where the $\boldsymbol{h}$ is the same as \ref{undetachable} and $\boldsymbol{\mu}$ is the Expected value of the probability distribution $p(x)$. So, with respect to the classical autoencoder this approach doesn't aim to have the same value to minimize the Loss, but instead to match the mean value of the distribution of the outcome. 
It is also straightforward to extend to the case of in-class distribution training, where, in that case, the minimum is reached when the expected value for each class distribution $\boldsymbol{\mu_j}$ is reached accordingly.

\section{Lower Bound and Reconstruction error with Random Sampling Training} \label{ReconstErrorRandom}
It's possible to find a lower bound of the Loss function and of the Reconstruction error of the same observation $ E[(f(g(x)) - x)^2]$ starting from the last findings.

\begin{align}
    \mathcal{L}_j(x, \theta, \phi)\bigg|_{\mathbb{E}[f(g(x,\theta),\phi)] = \boldsymbol{\mu_j}} &= \mathbb{E}_{x,y \sim p_j(x)}[(y - f(g(x, \theta),\phi))^2] \quad \forall j \nonumber \\
     \quad &= \mathbb{E}_{x,y \sim p_j(x)}[y^2 -2yf(g(x, \theta),\phi)) + f(g(x, \theta),\phi))^2] \quad \forall j \nonumber \\
     &\begin{multlined}
     \mathllap{\quad}  = \mathbb{E}_{y \sim p_j(x)}[y^2] - 2\mathbb{E}_{y \sim p_j(x)}[y]\mathbb{E}_{x \sim p_j(x)}[f(g(x, \theta),\phi)] + \nonumber \\
     + \mathbb{E}_{x \sim p_j(x)}[f(g(x, \theta),\phi))^2] \quad \forall j \nonumber
     \end{multlined} \nonumber\\
    \quad  &= \mathbb{E}_{y \sim p_j(x)}[y^2]-2\mu_j^2 +  \mathbb{E}_{x \sim p_j(x)}[f(g(x, \theta),\phi))^2] \quad \forall j \nonumber\\
\end{align}

Now, we remind the definition of variance $ Var(X) = E[X^2] - E[X]^2$, and we sum and subtruct $ E[f(g(x, \theta),\phi)]^2$

\begin{align}
    \mathcal{L}_j(x, \theta, \phi)\bigg|_{\mathbb{E}[f(g(x,\theta),\phi)] = \boldsymbol{\mu_j}} &= \mathbb{E}[y^2]-2\mu_j^2 +  \mathbb{E}[f(g(x, \theta),\phi))^2] \quad \forall j \\
    &\begin{multlined}\mathllap{\quad} = \mathbb{E}[y^2]-2\mu_j^2 +  \mathbb{E}[f(g(x, \theta),\phi))^2] + \mathbb{E}[f(g(x, \theta),\phi)]^2 + \\- \mathbb{E}[f(g(x, \theta),\phi)]^2  \quad \forall j \nonumber
    \end{multlined}\\
    \quad &= \mathbb{E}[y^2] -\mu_j^2   \mathbb{E}[f(g(x, \theta),\phi))^2]- \mathbb{E}[f(g(x, \theta),\phi)]^2  \quad \forall j \nonumber \\
    \quad &\geq Var(Y) + Var( f(g(X, \theta),\phi)) \quad \forall j \nonumber
\end{align}
Where the last inequality is the lower bound of the loss function for every class distribution. For the Full random sampling training, the inequality stays the same but the whole distribution corresponds to the whole dataset.

Now, to find the lower bound for the reconstruction error is enough to remember the definition of the variance and the finding $E[f(g(x))] = E[X]$ with less of notation, when the Loss is at the minimum. So, 
\begin{equation}
    \begin{split}
        Var(X) &= E[X^2] - E[X]^2 \\
        Var(f(g(X))) &= E[f(g(X))^2] - E[X]^2
    \end{split}
\end{equation}

Hence we have,
\begin{equation}
    \begin{split}
        Var(f(g(X))) &= E[f(g(X))^2] - E[X^2] - \sigma_x^{2} \\
        Var(f(g(X))) &= E[f(g(x))^2 - X^2] - \sigma_x^{2} \\
        Var(f(g(X))) &= E[(f(g(x)) - x)^2] +2E[xf(g(X))] - \sigma_x^2 \\
        E[(f(g(X)) - X)^2] &= \sigma_{f(g(X))}^2 + \sigma_x^2 - 2E[xf(g(X))] \\
        E[(f(g(X)) - X)^2] &= E[f(g(X))^2] + E[X^2] - 2E[xf(g(X))]
    \end{split}
\end{equation}
Where the last equation corresponds to the classical reconstruction error.

\section{Supplementary Materials}

\subsection{Models Architectures}
\begin{table}[ht]
    \centering
    \begin{tabular}{|c|c| c|}
 \hline
 Name & Architecture & Datasets\\ 
 \hline\hline
 Conv2d sizes & [32, 64, 128, 256, 512, 512, 256, 128, 64, 32] & \begin{tabular}{@{}c@{}} \quad \\ MNIST, \\ Fashion-MNIST\end{tabular} \\ 
 Kernel size & all 3x3 & \\
 Stride & all 1 & \\
 Padding & 0 & \\
 Batch Normalization & True & \\
 Activation Functions & LeakyRelu & \\
 Dropout & 0.2 & \\
 Last activation function & Sigmoid & \\
 
 \hline\hline

\begin{tabular}{@{}c@{}} \quad \\ Conv2d/ConvTranspose(c/ct) \\ 
sizes\end{tabular}   & \begin{tabular}{@{}c@{}} \quad \\ Encoder: [64, 64, 128, 128, 256](c), \\ Decoder[256(ct), 128(c), 128(ct), 64(c), 3(ct)]\end{tabular}   & \begin{tabular}{@{}c@{}} Caltech101 \\ CIFAR10\end{tabular} \\ 
\quad & \quad & \\
 Kernel size & Encoder: [3, 3, 3, 3, 3], Decoder: [3, 3, 3, 3, 3] & \\
 Stride & Encoder: [2, 1, 2, 1, 2], Decoder: [2, 1, 2,  1, 2] & \\
 Padding & Encoder: [1, 1, 1, 1, 1], Decoder: [1, 1, 1, 1, 1] & \\
 Output Padding & Decoder: [1, -, 1, -, 1] & \\
 Batch Normalization & True & \\
 Activation Functions & LeakyRelu & \\
 Dropout & 0.2 & \\
 Last activation function & Sigmoid & \\
    
\quad & \quad & \\
    \hline \hline

    \quad & \quad & \\
    Linear & [64, 8, 8, 64] & \\
    Activation functions & LeakyRelu &  BreastCancer\\
    Last Activation functions & Sigmoid & \\
    
\quad & \quad & \\
    \hline
\end{tabular}

    \caption{Model Architectures and Datasets}
    \label{tab:Tab1}
\end{table}

\end{appendices}


\bibliography{sn-bibliography}

\end{document}